\documentclass{article}

  \usepackage{arxiv}

  \usepackage[utf8]{inputenc} % allow utf-8 input
  \usepackage[T1]{fontenc}    % use 8-bit T1 fonts
  \usepackage[font=small,labelfont=bf]{caption}
  \usepackage{hyperref}       % hyperlinks
  \usepackage{url}            % simple URL typesetting
  \usepackage{booktabs}       % professional-quality tables
  \usepackage{amsmath}        % math environments (equation*, \text, \boldsymbol)
  \usepackage{amsfonts}       % blackboard math symbols
  \usepackage{amssymb}        % $\boxtimes$, $\boxdot$, $\square$ status markers
  \usepackage{nicefrac}       % compact symbols for 1/2, etc.
  \usepackage{microtype}      % microtypography
  \usepackage{lipsum}
  \usepackage{graphicx}
  \graphicspath{{./images/}{figures/}}
  \usepackage{longtable}      % multi-page tables (pandoc tables)
  \usepackage{array}          % extended column specs (pandoc tables)
  \usepackage{calc}           % length arithmetic (pandoc tables)
  \usepackage{etoolbox}       % conditionals used by pandoc output
  \usepackage{enumitem,geometry,titlesec,tcolorbox,xcolor}
  \usepackage[numbers]{natbib}

  \DeclareUnicodeCharacter{00A7}{\S}
  \DeclareUnicodeCharacter{00B1}{\ensuremath{\pm}}
  \DeclareUnicodeCharacter{00B2}{\ensuremath{^2}}
  \DeclareUnicodeCharacter{00B6}{\P}
  \DeclareUnicodeCharacter{00B7}{\ensuremath{\cdot}}
  \DeclareUnicodeCharacter{00D7}{\ensuremath{\times}}
  \DeclareUnicodeCharacter{0394}{\ensuremath{\Delta}}
  \DeclareUnicodeCharacter{03C4}{\ensuremath{\tau}}
  \DeclareUnicodeCharacter{2014}{\textemdash}
  \DeclareUnicodeCharacter{2019}{'}
  \DeclareUnicodeCharacter{2190}{\ensuremath{\leftarrow}}
  \DeclareUnicodeCharacter{2192}{\ensuremath{\rightarrow}}
  \DeclareUnicodeCharacter{21D2}{\ensuremath{\Rightarrow}}
  \DeclareUnicodeCharacter{21D4}{\ensuremath{\Leftrightarrow}}
  \DeclareUnicodeCharacter{2208}{\ensuremath{\in}}
  \DeclareUnicodeCharacter{2209}{\ensuremath{\notin}}
  \DeclareUnicodeCharacter{2212}{\ensuremath{-}}
  \DeclareUnicodeCharacter{222A}{\ensuremath{\cup}}
  \DeclareUnicodeCharacter{2248}{\ensuremath{\approx}}
  \DeclareUnicodeCharacter{2264}{\ensuremath{\le}}
  \DeclareUnicodeCharacter{2265}{\ensuremath{\ge}}
  \DeclareUnicodeCharacter{2282}{\ensuremath{\subset}}

\usepackage{xcolor}
\title{\textsc{TEDDY}: A Pediatric Foundation Model for Risk Forewarning from ICD-Coded Diagnostic Histories}

\author{
 Matthew Brady Neeley$^{\dagger}$ \\
  Baylor College of Medicine \\
  \texttt{matthew.neeley@bcm.edu} \\
  \And
 Jorge Botas$^{\dagger}$ \\
  Baylor College of Medicine \\
  \texttt{jorgeb@bcm.edu} \\
  \And
 Johnathan Jia \\
  Baylor College of Medicine \\
  \texttt{johnathan.jia@bcm.edu} \\
  \And
 Lin Yao \\
  Baylor College of Medicine \\
  \texttt{lin.yao@bcm.edu} \\
  \And
 Daniel Palacios \\
  Baylor College of Medicine \\
  \texttt{daniel.palacios@bcm.edu} \\
  \And
 Benjamin Choi \\
  Baylor College of Medicine \\
  \texttt{benjamin.choi@bcm.edu} \\
  \And 
 Zhandong Liu$^{*}$ \\
  Baylor College of Medicine \\
  \texttt{zhandonl@bcm.edu} \\
  \And
 Hyun-Hwan Jeong$^{*}$ \\
  Baylor College of Medicine \\
  \texttt{hyun-hwan.jeong@bcm.edu} \\
  \\[1em]
  \small $^{\dagger}$These authors contributed equally. \\
  \small $^{*}$Co-corresponding authors.
}

\begin{document}
\maketitle

\begin{abstract}
Pediatric electronic health records capture developmentally structured
clinical trajectories, yet their potential for generative
healthcare foundation models remains largely unexplored. Here we present
\textsc{TEDDY} (Temporal Event Decoder for Disease in Youth), a
1.84-million-parameter decoder transformer trained on
approximately 73 million ICD-10 diagnoses from 1.6 million children at a
single pediatric institution. \textsc{TEDDY} models longitudinal diagnosis
trajectories and visit timing. Predictions were made before visit codes were
revealed, limited to first occurrences, and evaluated against sex- and
age-matched controls.
Across 797 disease-onset prediction tasks spanning 16 ICD-10 chapters,
\textsc{TEDDY} achieved a median AUC of 72.0\%, outperforming same-data
DenseNet (50.0\%), CNN (57.2\%), RNN (60.1\%), and LSTM (62.7\%) baselines
on 96--99\% of tasks. Performance held
across sex and age and was strongest among lower-prevalence diagnoses; 202 of the
225 rarest conditions (90\%) had 95\% confidence intervals above
chance. Predictive signal remained detectable more than two years before first
recorded diagnosis, with median AUCs of 59.7\% in the unrestricted
analysis and 64.4\% in a fixed-cohort sensitivity analysis. In
asthma and attention-deficit/hyperactivity disorder benchmarks, AUCs were
79.3\% and 84.7\%, compared with 62.7\% and 71.7\% for the strongest
comparators, including a general-purpose language model three orders of
magnitude larger. Visit-timing
predictions had a 3.0-day mean absolute restricted mean
survival-time error over 365 days, although median and long-tail
return intervals remained miscalibrated. Together, these results establish
pediatric diagnostic histories as a substrate for compact generative models
supporting broad, rare-disease, and long-horizon risk forecasting without
population-scale data or billion-parameter models.
\end{abstract}

\clearpage % fixing intro cut off issue - DP 

\section{Introduction}

Pediatric patients differ from adults anatomically, physiologically, and in
disease presentation, yet most clinical prediction tools are trained on adult
data \citep{allegaert2008developmental}. Doses, normal vital-sign ranges, and
even the appearance of illness are age-dependent, so that sepsis in a
premature newborn looks nothing like sepsis in a teenager
\citep{wang2026transformative}. Because clinical
evidence is generated largely in adults, the protocols, practice guidelines,
and therapies derived from adult data cannot be applied directly to
children --- pediatric prescribing, for example, still leans heavily on
off-label use \citep{yuan2024off, antoon2023advancing}. The same limitation
now extends to the artificial-intelligence and machine-learning (AI/ML)
tools poised to reshape clinical care: only 17\% of FDA-approved AI/ML
devices are labelled for pediatric use \citep{hua2025lack, brewster2025us},
and of 181 public medical-imaging datasets, just 3.3\% are pediatric-only,
with a further 14.4\% mixing adult and pediatric data \citep{hua2025lack}.
Models trained mostly on adults can misread the physiology of children and
return unreliable predictions for the very patients they are meant to help.

The case for models built specifically for children rests on a simple fact:
the pediatric disease landscape is not a smaller version of the adult one.
Two classes of condition make the point. The first is developmentally
specific disease that has no adult counterpart --- congenital and genetic
disorders, complications of prematurity, and neurodevelopmental conditions
such as autism spectrum disorder --- which arise from and unfold along the
biology of early development \citep{van2022children, pradhan2026magnitude}.
The second is disease that clusters in childhood, such as asthma, allergy,
and epilepsy, where adult records are neither complete nor representative of
how the condition first presents and evolves in a child. In both classes, a
model trained on adult data inherits the wrong priors.

Rare disease sits at the sharp end of this second class and carries the
heaviest burden. Rare conditions are disproportionately pediatric in onset
\citep{wakap2020estimating}, and families spend more than six years on average
in search of a diagnosis
\citep{hendriksz2013rare, benitolozano2022diagnostic, yang2022national}, at
an estimated \$86{,}000--\$517{,}000 per patient --- a burden that reached
\$997 billion in the United States in 2019
\citep{everylife2023delayeddiagnosis}. Conventional decision-support tools
are not built to track illness that shifts with development
\citep{mullen2021timeliness, wang2026transformative}, so this diagnostic
odyssey persists even where the clues accumulate in the record.
Pediatric-specific, data-driven medicine offers a way forward. A child's
clinical trajectory is shaped by rapidly changing developmental physiology,
age-specific disease prevalence, preventive-care schedules, and close links
between maternal, infant, and early-life records, so a clinically useful
model must recognize not only what has already occurred in a child's record
but also what may emerge next as the child develops. Surfacing that trajectory at any
encounter that raises concern --- in the emergency department, in urgent
care, or at a specialty referral --- could prompt earlier testing, referral,
or admission before acute deterioration, bring appropriate intervention
forward, and cut the cost of a missed rare diagnosis. Yet open-vocabulary
forecasting across the broad range of pediatric diagnoses remains largely
unexplored.

The methods that might fill this gap were developed almost entirely in
adults, where transformers have been shown to learn disease trajectories from
longitudinal EHRs along two lines (Table~\ref{tab:sota_comparison}). The first adapts BERT-style models
--- BEHRT \citep{liBEHRTTransformerElectronic2020}, Med-BERT
\citep{rasmyMedBERTPretrainedContextualized2021}, CEHR-BERT
\citep{pangCEHRBERTIncorporatingTemporal2021}, and Hi-BEHRT
\citep{liHiBEHRTHierarchicalTransformer2023} --- which learn bidirectional
representations and then fine-tune a classifier for one task at a time, such
as heart-failure readmission or diabetes onset. Because they are
discriminative, these models do not generate future events and are often
tested on a handful of hand-picked conditions rather than across the whole
diagnostic range.

The second line is autoregressive and generative: it treats a patient's
history as a sequence of tokens and predicts the next medical event, and
when it will occur. \textit{Delphi} \citep{shmatkoLearningNaturalHistory2025}
adapts GPT-2 with a continuous age encoding and an exponential waiting-time
head, training on 0.4~million UK Biobank participants to predict the onset of
more than 1{,}000 ICD-10 diagnoses. \textit{ReClaim}
\citep{maFoundationModelsUnlock2026} scales the approach to 43.8~billion
events from over 200~million enrollees in a commercial-claims database, adds
procedures, medications, and cost, and reaches a mean AUC of 75.6\% across
more than 1{,}000 disease-onset tasks. Epic's \textit{Curiosity}
\citep{waxlerGenerativeMedicalEvent2025} shows the same models improve with
scale, training Qwen2 decoders of up to one billion parameters on
118~million patients from Epic Cosmos. Others --- Foresight
\citep{kraljevicForesightGenerativePretrained2024}, MOTOR
\citep{steinbergMOTORTimeEvent2024}, TransformEHR
\citep{yangTransformEHRTransformerBased2023}, and Life2vec
\citep{savcisensUsingSequencesLife2024} --- vary in architecture and data but
share the insight that autoregressive pretraining over clinical sequences
captures temporal structure that discriminative classifiers miss.

Two methodological pitfalls can make reported performance look better than it
is. The first is within-encounter leakage. EHR systems record diagnoses only
to the calendar day, so the several codes billed on one visit carry no order
among themselves. When a model is scored by predicting each code in turn
under a standard causal mask, a code can be predicted from its same-day
neighbours --- diagnoses that were not yet known when the visit began. If
asthma (J45) and pneumonia (J18) are billed the same day, such a setup can
let the model use the asthma code to predict the pneumonia code, inflating
apparent accuracy. How much this bites depends on how a model handles a
visit: approaches that generate or score a whole visit at once, or that
condition on a known reason for the encounter, are exposed differently. The
second pitfall is demographic confounding. If a sex-specific condition ---
say endometriosis or testicular torsion --- is scored against a mixed cohort,
most of the comparison group is of the opposite sex and trivially easy to
tell apart, so the model scores well on demographics alone rather than on any
clinical signal. Designing training and evaluation so that a model learns
real predictive structure, instead of exploiting either shortcut, is a
central methodological challenge.

These adult models also leave a population gap. Almost every generative EHR
model has been trained on adults or mixed-age cohorts. Notably,
\textit{Curiosity} excluded children outright, reasoning that pediatric care
differs too much from adult care, while naming pediatrics --- with its
maternal and infant links, developmental physiology, vaccination schedules,
and congenital timelines --- as a promising, unaddressed frontier. Claim-PT
\citep{zengPretrainedTransformerFramework2022} is, to our knowledge, the only
transformer pretrained on pediatric claims, but it is encoder-only and was
evaluated on specific tasks rather than open-vocabulary forecasting --- so
the generative paradigm has not yet been brought to children.

Pediatric records also differ in clinical structure. Childhood care is
organized around developmental transitions, preventive visits,
immunizations, developmental surveillance, and the early manifestations of
congenital and genetic disease \citep{aap2025periodicity, lipkin2020promoting, van2022children, pradhan2026magnitude}, and the same diagnosis can carry different
prevalence, implications, and expected timing at different ages \citep{wisk2025prevalence}. Well-child
and vaccination visits can dominate a longitudinal record without reflecting
disease progression, while maternal and infant histories that carry
important context are often held in separate records. A pediatric model must
therefore separate routine developmental care from evolving clinical risk
while staying sensitive to conditions that emerge within narrow developmental
windows.

Data availability compounds these obstacles. Public pediatric EHR data are
scarce --- large biobanks such as the UK Biobank \citep{sudlow2015ukbiobank} and MIMIC-IV \citep{johnson2023mimic}
are overwhelmingly adult --- and the pediatric records that do exist are
fragmented across systems \citep{forrest2014pedsnet}. Children with complex
conditions are seen only intermittently at major centers, while most of their
care happens with community providers whose records rarely reach those
centers, so any one institution sees only a truncated, unevenly sampled slice
of each child's story. The road to stronger performance in
adult EHR modelling has run through ever-larger datasets and models --- a road
largely closed to individual pediatric institutions, which must instead learn
from carefully curated, comparatively small data.

Meeting these challenges calls for a model whose predictions can be trusted
on the conditions that matter most --- which depends as much on how it is
evaluated as on how it is built. We present
\textsc{TEDDY} (Temporal Event Decoder for Disease in Youth), a decoder
transformer that reads a child's unfolding medical story and flags what may
be coming --- like an attentive pediatrician with perfect recall of every
similar child seen before. \textsc{TEDDY} is trained with a joint next-code
and time-to-event objective on 1.6~million patients and approximately
73~million ICD-10 diagnoses from the pediatric EHR of Texas Children's
Hospital (TCH).

We make three design choices in response: two close the shortcuts above, and
one keeps the task clinically meaningful. First, a visit-structure
representation brackets the codes of each same-day encounter and scores every
new diagnosis at the visit boundary --- before any of that day's codes are
revealed --- removing within-encounter leakage while keeping a standard causal
model. Second, we match each case to controls of the same sex and similar
age, so that per-condition accuracy reflects clinical signal rather than
demographics. Third, we score each condition only at its first occurrence, so
the task is anticipating a genuinely new diagnosis rather than re-predicting
one a child already carries. We then ask how far in advance each diagnosis can
be anticipated, reporting accuracy from the same day out to two or more years in advance.

Trained on the full ICD-10 vocabulary across all 22 chapters, \textsc{TEDDY}
reaches a median first-occurrence AUC of 0.72 on the 797 conditions
(validated across 16 disease-bearing chapters) that meet our thresholds. That accuracy holds across sex and age and is, if anything,
strongest for the rarest conditions. At just 1.84~million parameters it
outperforms four other sequence architectures and a general-purpose language
model three orders of magnitude larger. The signal is readable years before a diagnosis is recorded, and the model's timing estimates agree with observed timing at the decile level, though it miscalibrates both typical and long-tail return intervals.
% and its mistakes almost always stay within the child's own record rather than
% inventing a diagnosis --- recasting the usual hallucination worry as bounded,
% auditable errors of timing. 
Together these results give the first generative
EHR foundation model built for children, and show that careful trajectory
representation and rigorous evaluation can yield clinically meaningful
predictions even in the small-data, long-tailed world of childhood disease.

\begin{table}[!ht]
\centering
\caption{Comparison of transformer-based models for longitudinal medical event prediction. "Dx" = diagnoses, "Rx" = medications, "Px" = procedures. "Code Recurrence" indicates whether a model represents repeated occurrences of the same code in a patient's sequence rather than collapsing each code to its first appearance; it is a modeling property, distinct from this paper's first-occurrence evaluation protocol.}
\label{tab:sota_comparison}
\resizebox{\textwidth}{!}{%
\begin{tabular}{@{}lllllllll@{}}
\toprule
\textbf{Model} &
\textbf{Architecture} &
\textbf{Enc/Dec} &
\textbf{Pre-training Objective} &
\textbf{Data Source} &
\textbf{Population} &
\textbf{Event Types} &
\textbf{Model Size} &
\textbf{Code Recurrence} \\
\midrule
Delphi-2M \citep{shmatkoLearningNaturalHistory2025}
  & GPT-2
  & Decoder
  & Autoregressive + Time-to-event
  & UK Biobank (0.4M)
  & Adult % (37--73 yr)
  & Dx, death, lifestyle
  & 2.2M
  & No \\
ReClaim \citep{maFoundationModelsUnlock2026}
  & Qwen3
  & Decoder
  & Autoregressive + post-training
  & MarketScan Claims (200M)
  & Adult
  & Dx, Rx, Px, cost
  & 140M--1.7B
  & Yes \\
Curiosity \citep{waxlerGenerativeMedicalEvent2025}
  & Qwen2
  & Decoder
  & Autoregressive
  & Epic Cosmos (118M)
  & Adult % 18-120 years old
  & Dx, Rx, Px, labs
  & 62M--1B
  & Yes \\

BEHRT \citep{liBEHRTTransformerElectronic2020}
  & BERT
  & Encoder
  & Masked Language Modeling
  & CPRD (1.6M)
  & Adult (Mixed)
  & Dx
  & Not reported %? ($\sim$4M)
  & No \\

Med-BERT \citep{rasmyMedBERTPretrainedContextualized2021}
  & BERT
  & Encoder
  & Masked Language Modeling
  & Cerner (28M)
  & Adult
  & Dx
  & 17M
  & No \\

CEHR-BERT \citep{pangCEHRBERTIncorporatingTemporal2021}
  & BERT
  & Encoder
  & Masked Language Modeling + Visit Type Prediction
  & CUIMC-NYP (2.4M)
  & Adult
  & Dx, Rx, Px
  & 9M
  & No \\

Hi-BEHRT \citep{liHiBEHRTHierarchicalTransformer2023}
  & Hierarchical Transformer
  & Encoder
  & Bootstrap your Own Latent + Masked Language Modeling
  & CPRD (2.8M)
  & Adult
  & Dx, Rx, Px, labs
  & Not reported % ? ($\sim$2M)
  & No \\

TransformEHR \citep{yangTransformEHRTransformerBased2023}
  & Transformer
  & Encoder-Decoder
  & Autoregressive
  & VHA (6.5M)
  & Adult
  & Dx
  & Not reported
  & No \\

Foresight \citep{kraljevicForesightGenerativePretrained2024}
  & GPT-2
  & Decoder
  & Autoregressive
  & KCH/SLaM/MIMIC-III (0.8M)
  & Adult (Mixed)
  & Dx, Rx, Px, free text entities
  & Not reported %? ($\sim$55M)
  & Yes \\

MOTOR \citep{steinbergMOTORTimeEvent2024}
  & Transformer
  & Encoder
  & Time-to-event
  & MERATIVE/EHR-OMOP (56.6M)
  & Adult (Mixed)
  & Dx, Rx, Px, labs
  & 143M
  & No \\

life2vec \citep{savcisensUsingSequencesLife2024}
  & BERT
  & Encoder
  & Masked Language Modeling + Sequence Ordering Prediction
  & Danish National Registers (2.3M)
  & Adult
  & Dx, Rx, Px, lifestyle
  & 8.4M
  & No \\

Claim-PT \citep{zengPretrainedTransformerFramework2022}
  & Transformer
  & Encoder
  & Next Visit Prediction
  & PFK Claims (600k)
  & \textbf{Pediatric}
  & Dx, Rx, Px
  & Not reported
  & Yes \\
\midrule
\textbf{TEDDY (ours)}
  & GPT-2
  & Decoder
  & Autoregressive + time-to-event
  & TCH OMOP (1.6M)
  & \textbf{Pediatric}
  & Dx
  & 1.84M
  & \textbf{Yes} \\
\bottomrule
\end{tabular}%
}
\end{table}

\section*{Results}

% ———————————————————————
% 0. Incident-prediction framing (introduces AUC, umbrella phrase,
%    the three guards, and scope – ONCE).
% ———————————————————————
\textsc{TEDDY} is trained to read a child's accumulating record of hospital visits and predict the diagnoses that lie ahead — both at the next visit and over the years that follow. The central question for any such model is whether it can anticipate a diagnosis before it is recorded, so we built our evaluation around that question directly. For every condition, we identify the moment just before a child's first recorded diagnosis and ask whether TEDDY assigned that condition a higher risk for that child than for otherwise-similar children who never develop it. Aggregated across children, this yields a single measure of discrimination — the area under the ROC curve (AUC), where 0.5 is chance and 1.0 is perfect separation — which we report throughout.

\begin{figure}[htbp]
\centering
\includegraphics[width=\linewidth]{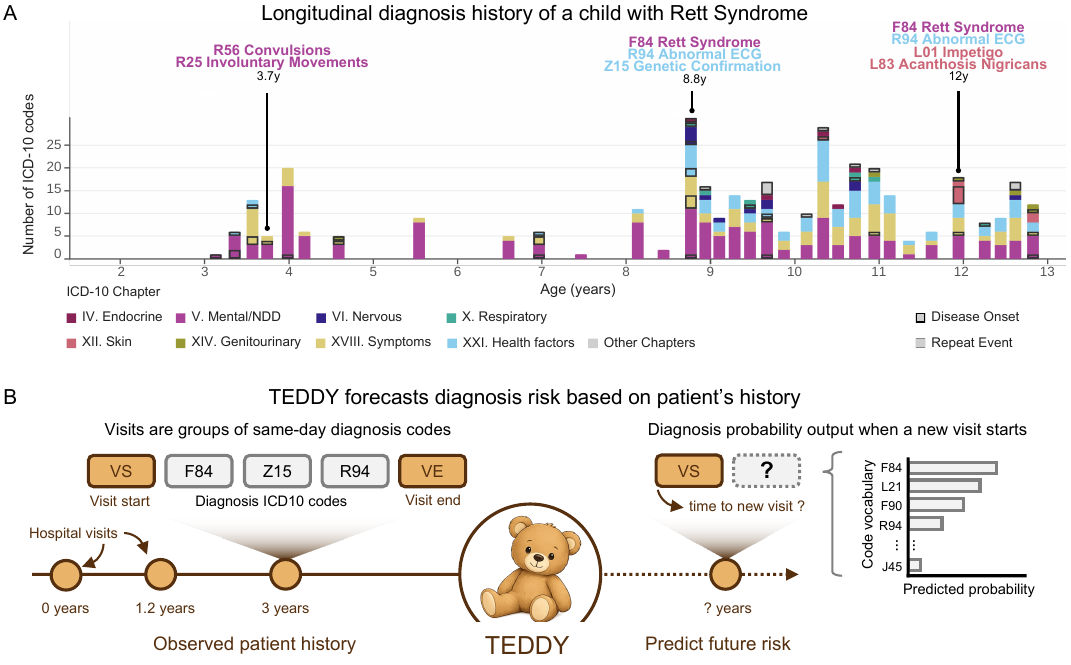}
\caption{\textbf{\textsc{TEDDY} forecasts pediatric diagnosis risk from
longitudinal visit histories.} \textbf{(A)} Example de-identified
diagnostic trajectory for a child with Rett syndrome. Each bar
summarizes ICD-10 diagnoses within a 60-day age window from approximately
2 to 13 years, stacked by ICD-10 chapter. Outlined segments mark first
recorded occurrences of diagnoses, whereas filled segments indicate
repeat events. The annotated milestones illustrate how early
neurodevelopmental, genetic-confirmation, neurologic, and later skin or
autonomic features appear along the same child’s longitudinal record.
\textbf{(B)} Overview of the visit-boundary prediction task. A child’s
history is represented as a sequence of hospital visits, where each visit
groups same-day ICD-10 diagnosis codes between a visit-start
(\textsc{VS}) and visit-end (\textsc{VE}) token. Given the observed visit
history, \textsc{TEDDY} predicts both the time to a future visit and the
diagnosis probability distribution over the closed ICD-10 vocabulary at
the start of that visit. Scoring at the visit boundary means predictions
are made before any same-day diagnosis codes are visible, preventing
within-encounter leakage while preserving the co-occurrence structure of
diagnoses recorded during the same clinical visit.}
\label{fig:teddy-overview}
\end{figure}

We score each condition only at
its first appearance in a child’s record, never at later repeats:
anticipating a genuinely new diagnosis is the task that matters
clinically, whereas a condition a child already carries is easy to predict
from its own past \citep{maFoundationModelsUnlock2026}. We ask
\textsc{TEDDY} for its prediction at the start of a visit, before it can
see any diagnosis entered that day, so it cannot quietly read codes that
happen to share the same calendar day and inflate its own accuracy
(Fig.~\ref{fig:teddy-overview}; Methods). We compare each child who develops a condition only with children of the same sex and similar age, preventing the model from achieving high performance simply by recognizing that a condition occurs predominantly in boys, girls, newborns, or other age-defined groups.
We report discrimination across the 16 ICD-10 chapters that capture
intrinsic childhood disease, spanning infectious conditions through
genitourinary disorders and including the perinatal and congenital
chapters. Chapters recording administrative or non-diagnostic information
— symptoms and abnormal findings, injuries and external causes,
health-service factors, special-purpose codes, and pregnancy — are
excluded. Unless we note otherwise, every AUC below is this same quantity:
first appearance, sex- and age-matched controls, computed on diagnosis histories the
model never saw during training.

% ———————————————————————
% 1. All-conditions view (Fig. 2)
% ———————————————————————
\subsection*{\textsc{TEDDY} anticipates incident diagnoses across the range of childhood illness}

\begin{figure}[htbp]
\centering
\includegraphics[width=\linewidth]{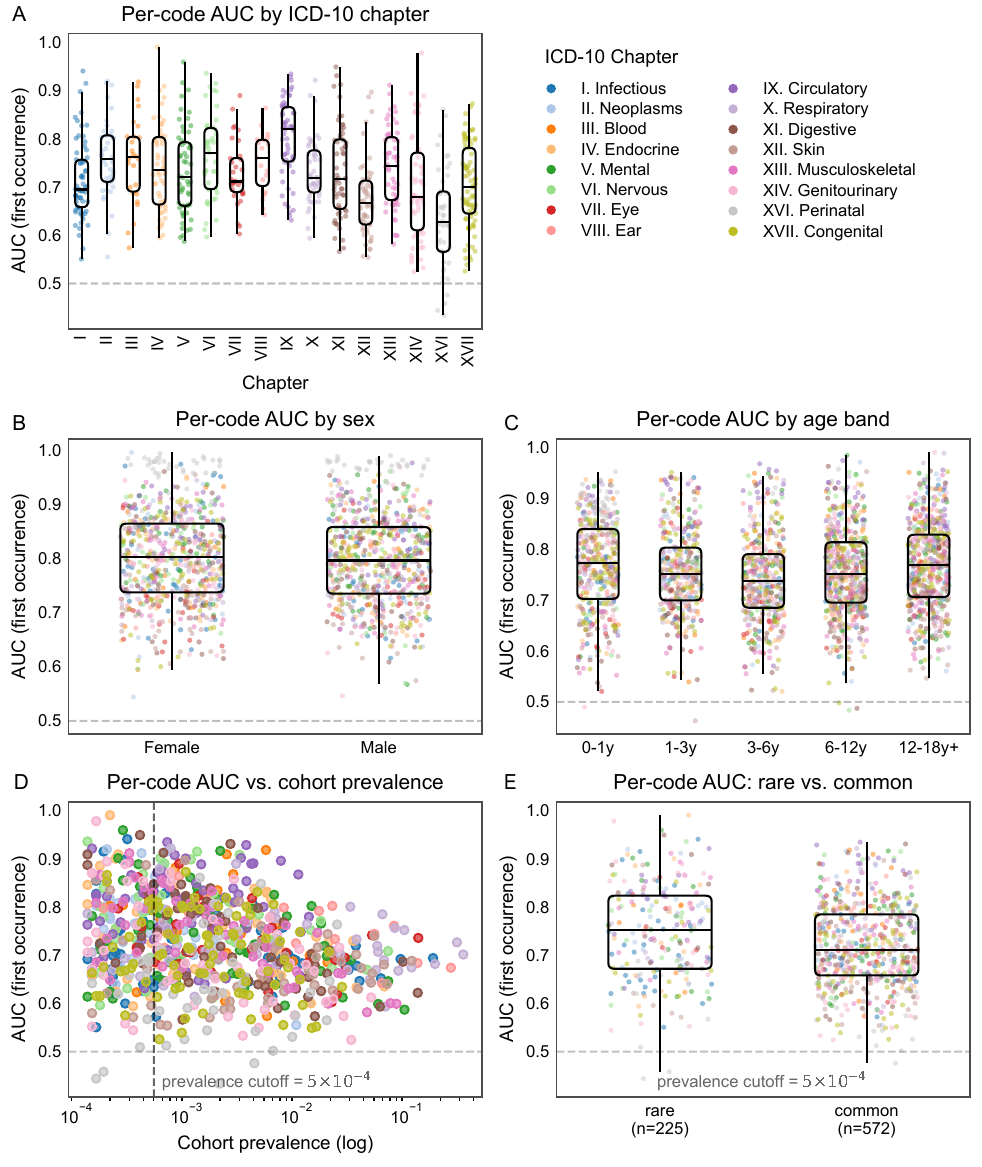}
\caption{\textbf{\textsc{TEDDY} anticipates
incident diagnoses across childhood illness, evenly across sex and age,
and most strongly for rare conditions.} Each point is one ICD-10 code;
the vertical axis is its first-occurrence AUC on the held-out test set,
measuring how well \textsc{TEDDY} ranks children who will receive that
diagnosis above children who never do, scored before the visit in which
the diagnosis first appears. Controls are matched to the sex and age
distribution of the children who develop each condition, so the AUC
reflects clinical signal rather than demographic separability.
\textbf{(A)} Per-code AUC by ICD-10 chapter. \textbf{(B)} Per-code AUC
stratified by sex. \textbf{(C)} Per-code AUC stratified by pediatric age
band. \textbf{(D)} Per-code AUC versus cohort prevalence on a log scale;
the dashed vertical line marks the rare/common cutoff of 5 in 10{,}000
children ($5\times10^{-4}$). \textbf{(E)} AUC distributions for the 225
rare and 572 common conditions defined by that cutoff. The dashed
horizontal line at AUC${=}0.5$ marks chance.}
\label{fig:percode}
\end{figure}

We first evaluated TEDDY across the full range of childhood illness, scoring every eligible condition rather than a hand-picked subset (Fig.\ref{fig:percode}). Across 797 conditions spanning the 16 ICD-10 chapters of intrinsic childhood disease, the median first-occurrence AUC was 0.72, and the per-condition AUC distribution remained above chance within every chapter (Fig.\ref{fig:percode}A).

\textsc{TEDDY} ranks risk similarly across sex and age. The distribution
of per-condition AUCs is comparable for girls and boys
(Fig.\ref{fig:percode}B) and across the five pediatric age bands from
infancy to late adolescence (Fig.\ref{fig:percode}C). Because each child
who develops a condition is compared only with controls of the same sex
and similar age, this consistency across sex and age strata is not explained by demographic
separability. Instead, it indicates that the model’s signal is broadly
distributed across the pediatric population rather than concentrated in a
demographically distinct subset of diagnoses.

\textsc{TEDDY}  performs particularly well where early recognition is often most
difficult. Its per-condition AUC is higher among lower-prevalence
diagnoses (Fig.~\ref{fig:percode}D), and the 225 rarest conditions,
each recorded in fewer than 5 of every 10{,}000 children,
($5\times10^{-4}$) — are anticipated at least as well as the 572 more
common conditions (Fig.\ref{fig:percode}E). A model that merely learned
base rates would be expected to degrade on the long tail of rare disease;
instead, \textsc{TEDDY} preserves discriminative signal for rare disease. This is not
only wider estimation scatter at low prevalence: with patient-level
bootstrap confidence intervals from 2{,}000 resamples, 202 of the 225
rarest conditions (90\%) have a 95\% CI lying entirely above chance
(Supplementary Fig.~\ref{fig:rare-ci}). The rare-code discrimination is
therefore a property of the model, not an artefact of estimating AUC from
few cases.

% ———————————————————————
% 2. Head-to-head benchmark (Fig. 3)
% ———————————————————————
% \subsection*{On asthma and Attention-Deficit/Hyperactivity
% Disorder, \textsc{TEDDY} leads every baseline tested}

\subsection*{\textsc{TEDDY} outperforms alternative architectures across
diagnoses and in focused clinical benchmarks}

Having established that \textsc{TEDDY} discriminates incident diagnoses
across the pediatric disease spectrum, we next asked whether this
performance could be recovered by alternative architectures trained on the
same records. We compared \textsc{TEDDY} with DenseNet, convolutional
neural network (CNN), recurrent neural network (RNN), and long short-term
memory (LSTM) baselines across the same 797 first-occurrence diagnosis
tasks. Each comparison was paired by ICD-10 code and used the same
visit-boundary, first-occurrence evaluation framework.

Across conditions, \textsc{TEDDY} showed a consistent upward shift in
per-code AUC relative to every same-data baseline
(Supplementary Fig.~\ref{fig:benchmark_all_codes}). The median
first-occurrence AUCs for the baseline models ranged from 0.500 for
DenseNet to 0.627 for LSTM, whereas \textsc{TEDDY} improved over each
comparator on 96--99\% of codes. The median paired improvement was largest
against DenseNet ($\Delta$AUC = 0.224) and remained substantial against
the strongest baseline, LSTM ($\Delta$AUC = 0.095). All four paired
comparisons were significant after Holm correction
(Supplementary Table~\ref{tab:paired_model_benchmark}). Thus,
\textsc{TEDDY}'s advantage is not driven by a small number of favorable
diagnoses and is not reproduced by simpler architectures trained on the
same pediatric trajectories.

\begin{figure}[htbp]
\centering
\includegraphics[width=\linewidth]{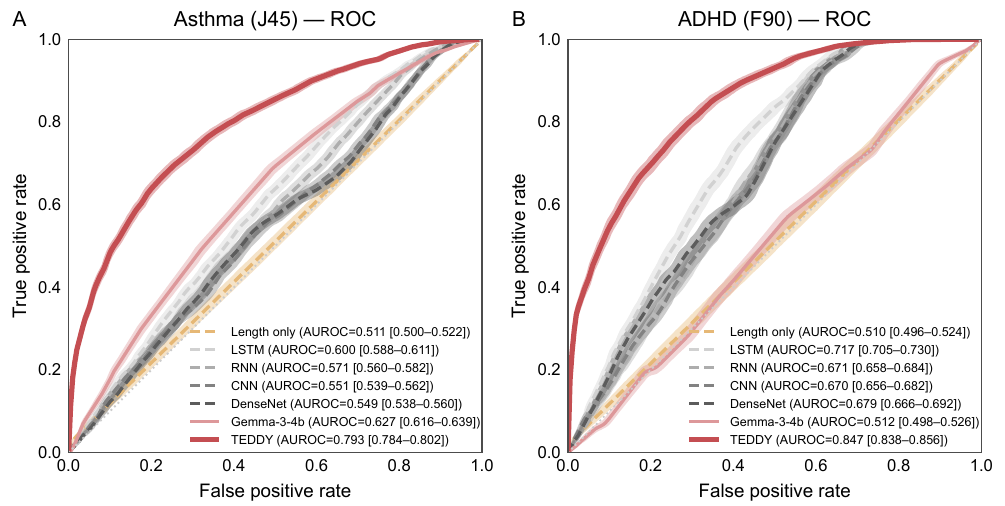}
\caption{\textbf{\textsc{TEDDY} outperforms same-data sequence models and
a larger general-purpose language model on incident asthma and ADHD
prediction.} Receiver operating characteristic curves compare models on
two pediatric diagnosis-onset tasks: \textbf{(A)} asthma (J45) and
\textbf{(B)} attention-deficit/hyperactivity disorder (ADHD; F90). For
each target, cases are children scored immediately before the visit
containing their first recorded diagnosis, and controls are children who
never receive that diagnosis. Control cutoffs are sampled to match the
case history-length distribution, preventing models from separating
groups simply by chart length. \textsc{TEDDY} is scored from a single
forward pass by reading the probability assigned to the target diagnosis
at the visit boundary, before any same-day codes are visible.
\textsc{TEDDY} achieves the highest AUROC for both targets: 0.793
(95\% CI 0.784–0.802) for asthma and 0.847 (0.838–0.856) for ADHD.
For asthma, the strongest non-\textsc{TEDDY} comparator is Gemma-3-4b
at 0.627 (0.616–0.639), and the strongest same-data sequence baseline
is the LSTM at 0.600 (0.588–0.611). For ADHD, the strongest
non-\textsc{TEDDY} comparator is the LSTM at 0.717 (0.705–0.730).
The length-only baseline remains near chance for both targets
(0.511 for asthma and 0.510 for ADHD). Shaded bands and bracketed
legend values denote 95\% patient-level bootstrap confidence intervals
from 2{,}000 resamples.}
\label{fig:benchmark}
\end{figure}

The broad per-code benchmark establishes the overall architecture-level
advantage, but it does not show the full operating characteristics for
individual clinically familiar outcomes. We therefore performed a focused
head-to-head analysis of two common and clinically important pediatric
diagnoses, asthma (J45) and attention-deficit/hyperactivity disorder
(ADHD; F90). This analysis additionally included a history-length-only
control and Gemma-3-4b, a general-purpose language model approximately
three orders of magnitude larger, prompted on the same patient histories
(Fig.~\ref{fig:benchmark}).

\textsc{TEDDY} ranked risk most accurately for both targets, reaching an
AUROC of 0.793 (95\% CI 0.784--0.802) for asthma and 0.847
(0.838--0.856) for ADHD. It exceeded the strongest non-\textsc{TEDDY}
comparator in each task with non-overlapping confidence intervals:
Gemma-3-4b reached 0.627 (0.616--0.639) for asthma, while the strongest
same-data sequence baseline for ADHD, the LSTM, reached 0.717
(0.705--0.730). Among the alternative architectures trained on the same
dataset, the LSTM was also the strongest asthma baseline at 0.600
(0.588--0.611). Confidence intervals here and below are 95\%
patient-level bootstrap intervals from 2{,}000 resamples and are shown as
shaded bands in Fig.~\ref{fig:benchmark} and Supplementary
Fig.~\ref{fig:benchmark_pr}.

The focused comparison was designed to minimize the influence of trivial
confounders. Every same-data sequence model saw the same children and
records, while cases and controls were matched on history length so that
no model could succeed merely by detecting that children with longer
charts tend to be sicker. A predictor using history length alone remained
near chance for both diagnoses (AUROC approximately 0.51). We also
included Gemma-3-4b~\citep{gemmateam2025gemma3technicalreport}, a
four-billion-parameter general-purpose language model prompted on the same
histories. It performed above chance for asthma but near chance for ADHD,
and remained well below \textsc{TEDDY} in both tasks. This comparison is
best interpreted as a scale control rather than a direct test of parameter
efficiency: a substantially larger general-purpose model without
pediatric-trajectory training did not approach a smaller model developed
specifically for longitudinal pediatric forecasting.

The same model ordering was observed under average precision, which
measures enrichment of children who subsequently receive the target
diagnosis (Supplementary Fig.~\ref{fig:benchmark_pr}).
\textsc{TEDDY} reached 0.805 (95\% CI 0.796--0.814) for asthma and
0.850 (0.840--0.860) for ADHD. For asthma, the strongest
non-\textsc{TEDDY} comparator was Gemma-3-4b at 0.594
(0.584--0.605), and the strongest same-data sequence baseline was the
LSTM at 0.572 (0.561--0.584). For ADHD, the strongest
non-\textsc{TEDDY} comparator was the LSTM at 0.658
(0.644--0.673). Together, the broad per-code comparison and focused
clinical benchmarks show that greater parameter count alone does not
recover the performance of a model trained specifically on pediatric
diagnosis trajectories.

% ———————————————————————
% 3. How-far-in-advance view (Fig. 4)
% ———————————————————————
\subsection*{\textsc{TEDDY}’s warning remains readable years in advance}

\begin{figure}[htbp]
\centering
\includegraphics[width=\linewidth]{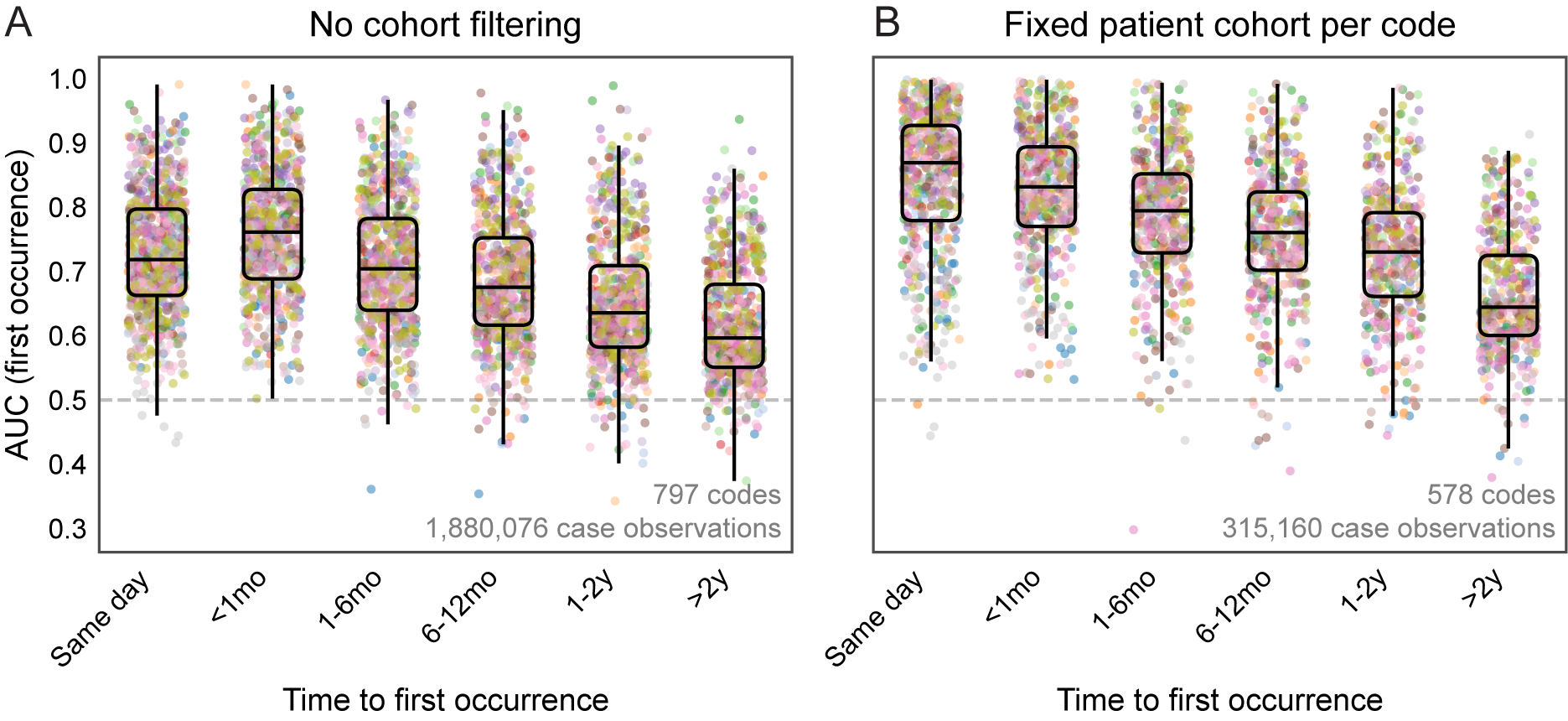}
\caption{\textbf{\textsc{TEDDY}’s diagnosis signal remains detectable
years before first recorded occurrence.} For each ICD-10 code, scored
visit-boundary positions are grouped by lead time to the next first
occurrence of the target diagnosis, and a separate first-occurrence AUC
is computed within each horizon band. Boxes summarize the distribution
of per-code AUCs across conditions; points are individual ICD-10 codes;
the dashed horizontal line marks chance (AUC${=}0.5$). \textbf{(A)} No
cohort filtering: every eligible first-occurrence case contributes to
its corresponding horizon band, so the contributing patients and codes
can vary across lead times. This analysis includes 797 codes and
$1{,}880{,}076$ first-occurrence case observations. Median AUC decreases
from 0.718 on the same day to 0.597 beyond two years, remaining above
chance across the full horizon range. \textbf{(B)} Fixed patient cohort
per code: for each code, the same case set is evaluated across all active
horizon bands, reducing composition shifts across lead times. This
analysis includes 578 codes and $315{,}160$ first-occurrence case
observations. Median AUC decreases from 0.869 on the same day to 0.644
beyond two years. Per-bin medians and 95% bootstrap confidence
intervals from 2{,}000 condition-level resamples are reported in
Supplementary Table~\ref{tab:future-horizon}.}
\label{fig:horizon}
\end{figure}

We define a warning as an interval before a condition's first recorded diagnosis during which TEDDY's risk estimate for that condition remains statistically discriminative between eventual cases and non-cases, a property of model output at a given lead time rather than a claim about a specific clinical alerting threshold. A warning is only useful if it arrives early enough to inform care. We therefore
asked not only whether \textsc{TEDDY} can anticipate a new diagnosis, but
how far before its first recorded occurrence that signal remains
detectable. For each condition, we grouped scored visit-boundary positions
by the time remaining until the target diagnosis first appeared, then
computed a separate first-occurrence AUC within each lead-time band
(Fig.\ref{fig:horizon}; Supplementary Table\ref{tab:future-horizon}).

In the unrestricted analysis, every eligible first-occurrence case
contributes to its corresponding horizon band. Across 797 conditions and
$1{,}880{,}076$ first-occurrence case observations, discrimination declined
gradually as the prediction horizon lengthened, but did not collapse to
chance. The median per-code AUC was 0.718 on the same day
(95\% CI 0.711–0.730), increased to 0.761 within the month before the first
diagnosis (0.750–0.771), and then decreased with longer lead times:
0.704 at 1–6 months (0.696–0.714), 0.675 at 6–12 months
(0.667–0.684), 0.636 at 1–2 years (0.624–0.646), and 0.597 beyond two
years (0.589–0.607). Thus, even when the diagnosis is still more than two
years away, the median condition remains above chance
(Fig.~\ref{fig:horizon}A).

Because the unrestricted analysis allows the contributing patients and
codes to vary by horizon, we repeated the analysis using a stricter fixed
patient cohort per code. In this sensitivity analysis, each code is
evaluated only on patients who are available as first-occurrence cases in
every active horizon band, so the decline is read on the same case set for
that code across lead times. This fixed-cohort analysis retained 578 codes
and $315{,}160$ first-occurrence case observations. The same temporal
pattern held, with higher absolute AUCs because the analysis is restricted
to codes and patients evaluable across all horizons: median AUC declined
from 0.869 on the same day (95\% CI 0.856–0.880) to 0.832 at
$<$1 month (0.821–0.843), 0.794 at 1–6 months (0.781–0.806), 0.760 at
6–12 months (0.750–0.771), 0.730 at 1–2 years (0.720–0.741), and
0.644 beyond two years (0.635–0.652)
(Fig.~\ref{fig:horizon}B).

Together, these analyses show that \textsc{TEDDY}’s signal fades with
distance from diagnosis but remains detectable years before the diagnosis
appears in the chart. That persistence argues against the model that
relies only on short-range links between neighboring visits. Instead,
\textsc{TEDDY} captures longer trajectory structure that could support a
forewarning role at routine pediatric visits, while making clear that
discrimination is strongest close to the eventual diagnosis and weaker at
multi-year horizons.

% ———————————————————————
% 4. Temporal calibration (Fig. 5)
% ———————————————————————
\subsection*{\textsc{TEDDY} estimates visit timing with small average error but long-tail drift}

\begin{figure}[htbp]
\centering
\includegraphics[width=\linewidth]{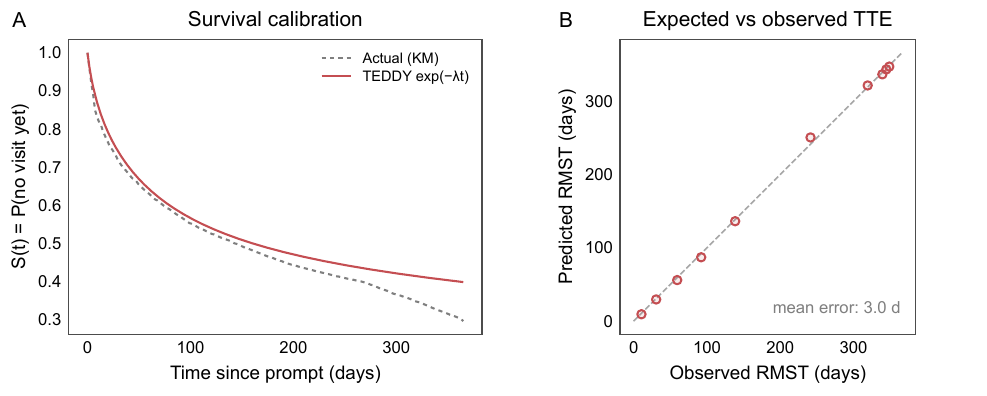}
\caption{\textbf{\textsc{TEDDY} estimates time to the next visit with
small average error but long-tail drift.} Predicted versus observed time
to the next visit, evaluated at $200{,}000$ sampled visit-boundary or
event-free positions from the held-out test set. \textbf{(A)} Survival
calibration: the empirical Kaplan–Meier curve shows the observed
fraction of children not yet back for a visit as time passes, compared
with \textsc{TEDDY}’s predicted exponential survival curve. The curves
agree most closely early after the prompt and diverge in the long tail;
by 365 days, the observed survival probability is 0.296 and the model
predicts 0.399. \textbf{(B)} Expected versus observed time to next
event: predicted and observed restricted mean survival times are aligned
across deciles of predicted RMST, with a mean absolute error of 3.0 days
over a 365-day horizon. Together, the panels show that \textsc{TEDDY}’s
timing head is accurate on average and patient-specific, while remaining
too simple to fully capture long inter-visit gaps.}
\label{fig:calibration}
\end{figure}

Beyond predicting a forthcoming diagnosis, \textsc{TEDDY} also estimates when a child will next be seen. We evaluated this estimate at the positions where the time-to-event head is trained to make it: visit boundaries and event-free intervals. This estimate is summarized as the restricted mean survival time (RMST), the expected time to the next event truncated at a fixed horizon. Across 200{,}000
 sampled positions from the held-out test set, predicted and observed RMST were closely aligned across deciles of predicted return time, with a mean absolute RMST error of 3.0 days over a 365-day horizon (Fig.~\ref{fig:calibration}B).

The timing estimate also varies meaningfully across children rather than
collapsing to a fixed follow-up interval. Among the $94{,}799$ uncensored
positions with an observed next visit, the model-implied and observed
gaps were correlated after log transformation (Pearson $r=0.54$ for
$\log(1+x)$-transformed predicted and observed times; Spearman
$\rho=0.57$ on the raw times). This indicates that \textsc{TEDDY} has
learned child-specific differences in return timing, not only the average
visit cadence of the cohort.

The fit is not perfect, and we report where it falls short. Among observed
return intervals, the observed median gap was 21.0 days, whereas the
model-implied exponential median was 38.8 days. The predicted survival
curve also drifts upward in the long tail: by one year, the observed
probability of no return visit is 0.296, whereas the model predicts 0.399
(Fig.~\ref{fig:calibration}A). This limitation reflects the deliberately
simple time head: a single exponential rate per position is easy to train
and interpret, but too rigid to capture the full spread of real
inter-visit intervals. Thus, \textsc{TEDDY}’s timing head is accurate on
average and patient-specific, but should not be interpreted as a fully
calibrated survival model.

\section*{Discussion}

We set out to ask whether a generative model could anticipate incident diagnoses across the pediatric disease spectrum 
— a setting where the data come from one institution, where most possible diagnoses are individually rare even though most visits involve just a handful of common diagnoses, and where tools built for adults transfer poorly. The credibility of
\textsc{TEDDY}'s numbers rests on the three guards built into its
evaluation. Scoring each diagnosis the instant a visit begins, before that
day's codes are visible, prevents the within-encounter leakage a standard
model would otherwise exploit. Scoring only first occurrences focuses the evaluation on the clinically important task of anticipating a newly recorded diagnosis -- anticipating a new diagnosis
--- rather than re-predicting a condition a child already carries. And
matching each case to controls of the same sex and similar age limits
demographic composition from passing as clinical signal, a confound that
inflates accuracy for any condition concentrated in one sex or age range.
Each safeguard addresses a distinct route through which EHR forecasting performance can appear stronger. 
Together, they make \textsc{TEDDY}'s discrimination a more
stringent estimate of predictive structure, rather than an artifact of
same-visit information leakage, prior diagnosis documentation, or
demographic imbalance.

Read that way, two features of the results carry clinical weight.
\textsc{TEDDY}'s discrimination is broadly consistent across females and
males and across the pediatric age range, indicating that its predictive
signal is not confined to a demographically distinct subset of childhood
illness. Moreover, contrary to the degradation often expected on
long-tailed clinical data, median discrimination was highest among the rarest conditions 
rather than approaching chance as prevalence falls. This is
particularly important in pediatrics: rare congenital and genetic
disorders account for a substantial share of the diagnostic landscape and
are among the conditions for which recognition is most often delayed
\citep{hendriksz2013rare}. 
That the signal remains detectable years before a diagnosis is recorded, 
rather than emerging only shortly before clinical recognition, 
motivates future evaluation of whether such signals could support risk review during routine pediatric encounters.

These results came from a model of only 1.84 million parameters trained on
a single institution's records. In the head-to-head view it outperformed four
other sequence architectures trained on the same data and a
general-purpose language model three orders of magnitude larger. The usual
route to stronger EHR models has been ever-larger adult datasets and
ever-larger models
\citep{shmatkoLearningNaturalHistory2025, maFoundationModelsUnlock2026, waxlerGenerativeMedicalEvent2025},
a path largely closed to individual pediatric centers. 
Our results suggest that, in this setting, faithfully representing a
child's unfolding clinical trajectory may matter more than model size,
which is encouraging for pediatric institutions that lack access to
population-scale training corpora.

% A well-known concern about generative models is that they hallucinate ---
% producing fluent output that the input does not support. For a model over
% a closed vocabulary the concern needs restating: every prediction is a
% real code by construction, so the question is whether a prediction falls
% within the child's own documented record. Out-of-history predictions ---
% the closest this kind of model comes to inventing a diagnosis --- were
% vanishingly rare, and when \textsc{TEDDY} erred it almost always named a
% real condition from the child's own record at the wrong time rather than
% one the child never receives. That recasts the worry as one of bounded,
% auditable errors of timing and detail --- far more compatible with
% clinical review than open-ended fabrication.

\textsc{TEDDY}'s sense of timing is calibrated where it is trained to be.
Its predicted gap to the next visit tracked the observed gap with small
average error, confirming that it learned child-specific visit rates
rather than a single rate for everyone. The estimate over predicts the
typical interval and departs from the observed long tail --- the expected
price of summarizing a skewed distribution of waiting times with a
one-number rate --- and we report this as a known limitation rather than
smoothing over it.

Several limitations bound these conclusions. The records come from a
single institution. Although TCH's 1.6 million pediatric patients span a
broad clinical population, coding practices, documentation patterns, and
case mix vary across health systems, and we have not established whether
\textsc{TEDDY}'s discrimination generalizes across those differences. The
vocabulary is limited to ICD-10 diagnoses, so the model does not incorporate
other structured EHR data, including vital signs, growth and BMI
percentiles, laboratory values, imaging, procedures, or medications, which
may provide additional predictive signal.

Three extensions follow naturally. The first is to widen the input beyond diagnoses, to measurements, medications, and procedures, and to test \textsc{TEDDY} across institutions and common data models. The second is intergenerational: the same model can be fine-tuned on mother-child pairs, with maternal prenatal history prepended to the child's record, to ask whether maternal context gives calibrated predictive signal for the child. We have curated and held out a pretrain-disjoint dyad cohort for exactly this purpose. The third concerns recurrent critical events, such as repeated emergency presentations or relapses of an existing diagnosis, which we have not analyzed in depth here because a model can learn to anticipate a patient's own recurring history rather than a risk pattern that generalizes, inflating apparent accuracy. Evaluation designs that separate first-occurrence prediction from recurrence prediction, or that limit the model's access to a patient's own prior instances of the same event, may reveal whether strong performance on recurrent events reflects genuine early signal or simple pattern repetition. A prospective, multi-site, multimodal pediatric trajectory model is the goal toward which this work is a first step.

% \section*{Conclusion}

% \textsc{TEDDY} is a generative EHR foundation model built for children and
% evaluated under a leakage-free, demographically matched, first-occurrence
% protocol. On that footing it anticipates incident diagnoses across the
% pediatric ICD-10 range --- evenly across sex and age, most sharply for the
% rarest conditions, and years before they are recorded. At just
% 1.84~million parameters it outperforms sequence baselines trained on the
% same data and a far larger language model. Its mistakes stay within each
% child's own record, recasting the hallucination worry as bounded errors of
% timing. Extending the input across measurement channels and lab values, institutions, and
% mother--child trajectories is the path toward clinically actionable prediction.

\section*{Conclusion}

\textsc{TEDDY} is a generative EHR foundation model built for children and
evaluated under a leakage-free, demographically matched, first-occurrence
protocol. On that footing, it anticipates incident diagnoses across the
pediatric ICD-10 range --- evenly across sex and age, most sharply for the
rarest conditions, and years before they are recorded. At just 1.84 million
parameters, it outperforms sequence baselines trained on the same data and
a far larger language model. Extending the input across measurement
channels and laboratory values, institutions, and mother--child
trajectories is the path toward clinically actionable prediction.

% =====================================================================
% Methods --- Teddy Foundation Model (manuscript draft)
% Companion to docs/manuscript_outline.tex
%
% Drafting status legend (per subsection):
%   [DRAFT]    Brady has written prose; awaiting Claude polish
%   [POLISH]   Claude has polished; awaiting Brady review
%   [FINAL]    Brady-approved; ready for manuscript
%   [SCAFFOLD] Claude has dropped key points / equations / references
%              for Brady to expand into prose
%   [TODO]     Not yet started
% =====================================================================

\section*{Methods}

% ---------------------------------------------------------------------
% 1. Data and cohort                                            [POLISH]
% ---------------------------------------------------------------------
\subsection*{Data and cohort}

We assembled a single-site pediatric electronic health record corpus from
Texas Children's Hospital (TCH), comprising condition occurrences and
patient demographics as of a 2026-01-22 data snapshot. We retained patients with at least
three distinct ICD coded events and age $\leq 25$ years, after within-encounter
de-duplication (a diagnosis code recorded more than once for the same
patient on the same calendar day is collapsed to a single event).
Mother--infant dyads were identified using linkages recorded in the
EHR, which associates each maternal delivery encounter with the
corresponding newborn encounter at the time of birth. All dyad
participants (mothers and infants; 81{,}847 patients) were excluded
from the main pretraining corpus and held out as a patient-level
disjoint cohort, so that any inter-generational transfer experiment on
these dyads would be free of pretraining leakage (held as future work;
see Discussion).

The final pretraining corpus contains 1{,}602{,}216 patients
and $\approx$73 million ICD-10 condition events, drawn from a
closed vocabulary of 1{,}849 tokens: 1{,}841 ICD-10-derived
codes plus eight special tokens (\texttt{PADDING}, \texttt{BIRTH},
\texttt{NO\_EVENT}, \texttt{SEX\_FEMALE}, \texttt{SEX\_MALE},
\texttt{SEX\_UNKNOWN}, \texttt{VISIT\_START}, \texttt{VISIT\_END}).
ICD codes are truncated to chapter and subcategory (decimal removed)
prior to tokenization. At training time, to make prediction targets well-defined during
long event-free intervals, we insert \texttt{NO\_EVENT} tokens by
\emph{gap-fill}: between any two consecutive real events, a cursor
advances forward by an age-appropriate interval drawn from the American Academy of Pediatrics (AAP)
well-child periodicity schedule
\citep{aap2025periodicity} (Table~\ref{tab:aap-cadence}), emitting a
\texttt{NO\_EVENT} token at each step until reaching the next real
event. Real events are never displaced.
Finally, we bracket each clinical visit: every maximal run of condition
codes that a patient receives on the same calendar day is wrapped between
a \texttt{VISIT\_START} and a \texttt{VISIT\_END} token (both carrying
that day's age). Bracketing is applied after \texttt{NO\_EVENT} gap-fill,
so \texttt{BIRTH} and \texttt{NO\_EVENT} tokens remain between visits
rather than inside them. We split patients
90/5/5 into train, validation, and test partitions and use the same patient-level
split for every analysis in the paper unless explicitly noted.

\begin{table}[h]
\centering
\small
\caption{AAP well-child cadence used for
\texttt{NO\_EVENT} gap-fill. Bucket boundaries are upper-inclusive on
age in days, so a real-event gap that begins in one bucket may produce
\texttt{NO\_EVENT} tokens at intervals from later buckets as the cursor
ages.}
\label{tab:aap-cadence}
\begin{tabular}{lll}
\toprule
Age (days) & Age (clinical) & Interval (days) \\
\midrule
$\leq 30$        & $\leq 1$ month       & 15  \\
31--365          & 1--12 months         & 45  \\
366--1{,}095     & 1--3 years           & 135 \\
1{,}096--4{,}380 & 3--12 years          & 270 \\
4{,}381--6{,}570 & 12--18 years         & 365 \\
$>6{,}570$       & $>18$ years          & 365 \\
\bottomrule
\end{tabular}
\end{table}

Every record is de-identified before modeling: a patient is represented
  solely by an integer study identifier, an ordered sequence of condition
  tokens, and, for each token, the patient's age in whole days at the time
  of the event. Temporal information is stored exclusively as an age offset
  relative to birth (day $0$), so no calendar dates, admission or service
  times, ages in years, or other date-shiftable quantities are retained.
  The pipeline ingests only the patient identifier, the ICD-10 condition
  code, and this age-in-days offset; no names, medical record numbers,
  free-text notes, provider identifiers, or geographic fields are read at
  any stage. The data-snapshot date is used only to apply the age-eligibility
  cutoff and is not carried into the modeling corpus. Consequently the
  tokenized dataset consists entirely of surrogate identifiers, categorical
  condition codes, and integer day-offsets, and contains none of the
  protected health identifiers enumerated under the HIPAA Safe Harbor
  provision.

% =====================================================================
% End draft prose --- delete the above block once Brady has written §1.
% =====================================================================

% =====================================================================
% Subsections 2--14 will be drafted iteratively.
% =====================================================================

% ---------------------------------------------------------------------
% 2. Tokenization and trajectory representation                  [POLISH]
% ---------------------------------------------------------------------
\subsection*{Tokenization and trajectory representation}

Each patient is represented as a chronologically ordered sequence of
(token, age) pairs drawn from the closed 1{,}849-token vocabulary
described in the Data and cohort section, with ages stored as integer days from birth. The
trajectory begins with a synthetic \texttt{BIRTH} token at age 0,
followed by the patient's \texttt{SEX\_*} token (also at age 0) and
any condition codes recorded on the day of birth; subsequent condition
codes appear in age order, and \texttt{NO\_EVENT} tokens are inserted
between consecutive real events according to the AAP gap-fill cadence
(Table~\ref{tab:aap-cadence}).
Consecutive condition codes recorded on the same day are then grouped
into a \emph{visit}: each such same-day run is delimited by a
\texttt{VISIT\_START} and \texttt{VISIT\_END} token pair, while
\texttt{BIRTH} and \texttt{NO\_EVENT} tokens pass through unbracketed
between visits. Bracket tokens inherit the age of the visit they enclose.

Each patient is a single training example. We use a context window of
2{,}048 tokens; the bracketed trajectories have a median length
of 59 tokens (p99 731), and the 0.23\% of patients whose trajectory
exceeds the context window are left-cropped to retain only their most
recent 2{,}048 tokens. Shorter sequences are right-padded
with a dedicated \texttt{PADDING} token, and padded positions are
ignored by both the loss and the attention mask. Events that share an
integer age (i.e.\ same-day events) are randomly reordered each epoch
as a stochastic data-augmentation step; this does not change which
tokens are visible at any prediction position.

% ---------------------------------------------------------------------
% 3. Model architecture                                          [POLISH]
% ---------------------------------------------------------------------
\subsection*{Model architecture}

  \textbf{Architecture.} \textsc{TEDDY} is a decoder-only transformer
  (8 pre-norm layers, 8 attention heads, embedding dimension $n_{\mathrm{embd}}=128$;
  \citealp{vaswani2017attention,radford2018gpt}) with weight-tied input
  and output embeddings, yielding $1{,}839{,}488$ trainable parameters
  ($\approx$1.84M). Each input position carries two
  embeddings that are summed before the first transformer block: a learned
  token embedding $\mathbf{e}_{\mathrm{tok}}(x_t) \in \mathbb{R}^{128}$
  indexed by the token id $x_t \in \{0, \ldots, 1{,}848\}$, and an
  \emph{age encoding} $\mathbf{e}_{\mathrm{age}}(a_t) \in \mathbb{R}^{128}$
  that maps the patient's age in days $a_t$ to a sinusoidal feature vector
  (period bank with maximum wavelength $10^{4}$, applied to age in years
  $a_t/365.25$) followed by a bias-free linear projection. We use no
  learned positional embeddings: temporal ordering enters the model
  exclusively through this age channel and through a standard triangular
  (causal) attention mask, under which position $t$ attends to all
  positions $s \leq t$. Each transformer block consists of LayerNorm,
  multi-head causal self-attention, residual connection, LayerNorm, and a
  GELU MLP with 4$\times$ expansion. We use dropout $p=0.1$ throughout and
  a final LayerNorm before the output projection.

  \textbf{Output and loss.} Our training objective builds on the
  Delphi-2M framework for joint token-and-time modeling of disease
  trajectories \citep{shmatkoLearningNaturalHistory2025}, which our work extends in
  several directions described in the following sections. A single linear head (weight-tied to the input embedding)
  projects the final hidden state at position $t$ to logits
  $\mathbf{z}_t \in \mathbb{R}^{1849}$. These logits parameterise two
  things simultaneously: (i) the next-token distribution, used in the
  categorical cross-entropy loss $\mathcal{L}_{\mathrm{CE}}$ over the
  predicted token; and (ii) the rate
  $\lambda_t = \exp(\mathrm{logsumexp}(\mathbf{z}_t))$ of an exponential
  time-to-next-\emph{visit} distribution, used in a negative
  log-likelihood loss $\mathcal{L}_{\Delta t}$ over the interval
  $\Delta t_t = a_{t+1} - a_t$ to the next visit boundary. Because visits
  are explicitly bracketed (the Tokenization and trajectory representation section), the model predicts elapsed time only at
  positions whose target begins a new visit or a \texttt{NO\_EVENT}
  interval --- that is, the time clock advances per visit transition
  rather than at every within-visit code --- whereas the categorical
  loss applies at every predictable position.

  The \texttt{NO\_EVENT} tokens are inserted on a fixed well-child cadence
  rather than observed at true encounters, so they mark points at which a
  visit had \emph{not yet} occurred: right-censored observations of the
  latent time-to-next-visit rather than events. We therefore treat every
  \texttt{NO\_EVENT} target as right-censored in the time loss. Such
  positions contribute the survival term
  $-\log S(\Delta t_t) = \lambda_t\,\Delta t_t$ rather than the full event
  density $-\log f(\Delta t_t) = -\log\lambda_t + \lambda_t\,\Delta t_t$,
  while positions whose target is a genuine visit boundary contribute the
  event density as usual. This censoring applies only to the time loss;
  \texttt{NO\_EVENT} remains an ordinary categorical target in
  $\mathcal{L}_{\mathrm{CE}}$. Because \texttt{NO\_EVENT} accounts for
  roughly half of inter-visit positions (up to $\sim$80\% in infancy),
  charging the full event density at these positions otherwise biases the
  learned rate toward the synthetic cadence grid. We train against the
  convex combination
  $\mathcal{L} = \mathcal{L}_{\mathrm{CE}} + \mathcal{L}_{\Delta t}$.
  Context-only tokens (\texttt{PADDING}, \texttt{BIRTH}, \texttt{SEX\_*})
  are excluded from the prediction targets and the loss; \texttt{NO\_EVENT},
  the visit-bracket tokens (\texttt{VISIT\_START}, \texttt{VISIT\_END}),
  and ICD codes are all predictable.

  % ---------------------------------------------------------------------
  % 4. Training                                                    [POLISH]
  % ---------------------------------------------------------------------
  \subsection*{Training}

  We optimise the composite loss
  $\mathcal{L} = \mathcal{L}_{\mathrm{CE}} + \mathcal{L}_{\Delta t}$ with
  AdamW \citep{loshchilov2019decoupled} and a cosine learning-rate
  schedule peaking at $6\times 10^{-4}$ after a linear warmup over the
  first 10\% of training iterations, for five epochs. We train in
  bfloat16. Padded positions and context-only tokens
  (\texttt{PADDING}, \texttt{BIRTH}, \texttt{SEX\_*}) are masked from
  attention and excluded from prediction targets; \texttt{NO\_EVENT} and
  the visit-bracket tokens remain predictable under
  $\mathcal{L}_{\mathrm{CE}}$, while \texttt{NO\_EVENT} targets are treated
  as right-censored in the time loss $\mathcal{L}_{\Delta t}$. The headline
  model was trained data-parallel across eight NVIDIA A100 GPUs, giving an
  effective batch of 32 sequences per
  optimizer step.
% VERIFY (§4 hardware/batch): GPU count, effective batch (8 x 4 x 1 = 32),
% and the per-epoch token figure are from the origin/main run notes, not
% from the local config. Confirm against the actual headline training job
% before submission (batch_size, gradient_accumulation_steps, world size).

% ---------------------------------------------------------------------
% 5. Per-code AUC: estimation and demographic matching           [POLISH]
% ---------------------------------------------------------------------
\subsection*{Per-code AUC: estimation and demographic matching}

For each of the $1{,}841$ ICD codes $c$ (of the $1{,}849$-token
vocabulary), we estimate a per-code AUC with the Mann--Whitney $U$
statistic comparing the model's logit for $c$ between cases and
controls. The logit is always read at a \texttt{VISIT\_START} position
--- the model's estimate, on entering a visit, of the codes that visit
will contain. Reading the score there, rather than from a neighbouring
same-day code, is what makes the estimate leakage-free (the Tokenization and trajectory representation section): because the
codes recorded on a single calendar day carry no inherent order, scoring
at \texttt{VISIT\_START} takes the prediction before any of that day's
codes have been supplied as input. A \emph{case} for $c$ is a
\texttt{VISIT\_START} opening a visit in which $c$ is recorded; a
\emph{control} is a \texttt{VISIT\_START} from a patient who never has
code $c$ recorded anywhere in their trajectory. Each patient contributes
a single such position to the estimate. Codes with fewer than
20 cases or 20 controls are excluded --- the case
threshold is set low to retain rarer codes.
% VERIFY (§5 thresholds): compute_per_code_auc default is min_cases=20,
% min_controls=20 (scripts/evaluate.py on the local branch). Confirm the
% headline eval lowered min_cases to 10; if it used 20, change the number
% and drop the "set low to retain rarer codes" clause. We further restrict the reported codes to the
We retain sixteen ICD-10 chapters that capture intrinsic childhood disease
and exclude chapters that record administrative or non-diagnostic
information:
symptoms and abnormal findings (\textsc{R}); injuries and external causes
(\textsc{S}--\textsc{T}, \textsc{V}--\textsc{Y}); factors influencing
health status and contact with services (\textsc{Z}); special-purpose
codes (\textsc{U}); and pregnancy, childbirth, and the puerperium
(\textsc{O}). The codes that pass the case threshold within these
chapters are the conditions scored throughout (Results). We report
first-occurrence AUC (the
case restricted to the first visit in which $c$ appears), with a 95\% confidence interval.

The discrimination confidence intervals for both the per-code AUCs here (Fig.~\ref{fig:percode}, Supplementary Fig.~\ref{fig:rare-ci}) and the head-to-head AUROC and average-precision intervals of the model benchmark (Fig.~\ref{fig:benchmark}, Supplementary Fig.~\ref{fig:benchmark_pr}) are computed by the same patient-level nonparametric bootstrap: the units being scored are
resampled with replacement $2{,}000$ times, the same estimator is
recomputed on each replicate, and the $2.5$th and $97.5$th percentiles of
the resulting distribution are taken as the interval. The two analyses
differ only in what one unit is and which estimator is recomputed. For
the per-code AUC, each patient contributes a single scored position, so
resampling the case and control rows is equivalent to resampling
patients, and each replicate recomputes the sex- and age-matched weighted
Mann--Whitney AUC defined below; for the benchmark, the unit is a cohort
patient's $(\text{label},\text{score})$ pair and each replicate recomputes
the unweighted AUROC and average precision (and, for the curve bands, the
ROC and precision--recall curves interpolated onto a common grid).
Resampling the matched estimator itself, rather than using the analytic
DeLong variance for the unweighted statistic, propagates the uncertainty
of the demographic reweighting into the per-code intervals; in both
analyses the bootstrap centre reproduces the reported point estimate (to
within $10^{-4}$ for every per-code AUC), so each interval is anchored to
the value it accompanies. The future-horizon median confidence intervals (§6; Supplementary Table~\ref{tab:future-horizon}) use a different unit and are not part of this patient-level procedure: because the reported quantity is the median AUC across conditions within a horizon band, each interval is a condition-level bootstrap, resampling the set of per-code AUC values with replacement $2{,}000$ times and taking the $2.5$th and $97.5$th percentiles of the resulting medians.

We examine the per-code intervals in particular for the rare stratum, to
separate genuine rare-code signal from the wider estimation scatter
expected at low prevalence: $202$ of the $225$ rarest conditions ($90\%$)
have a first-occurrence interval lying entirely above chance
(Supplementary Fig.~\ref{fig:rare-ci}).

To ensure that cases and controls are demographically comparable in the
pooled estimate, we reweight controls to match the joint sex and
age-band distribution of the cases for each code. Sex is encoded as
female, male, or unknown (unknowns are matched only to unknowns). Each
control receives a non-negative weight $w = (n^{\mathrm{case}}_{sb} /
n^{\mathrm{case}}) / n^{\mathrm{ctrl}}_{sb}$ within its (sex, band) cell
$sb$; strata in which cases exist but no control is available are dropped
and their case mass reported as a coverage deficit. Age bands follow a
coarsen-then-fall-back cascade: we begin with a fine 15-band pediatric
schema (0--1mo, 1--3mo, 3--6mo, 6--12mo, 1--1.5y, 1.5--2y, 2--3y, 3--4y,
4--6y, 6--8y, 8--10y, 10--12y, 12--15y, 15--18y, 18y+) and, whenever
case-coverage falls below 0.5, fall back first to the 5-band AAP scheme
(0--1y, 1--3y, 3--6y, 6--12y, 12--18y+), then to a 3-band coarse scheme
(0--3y, 3--12y, 12--18y+), and finally to sex-only matching. For each
code we report the Kish effective sample size
$n_{\mathrm{eff}} = (\sum_i w_i)^2 / \sum_i w_i^2$, which equals the
control count under uniform weights and shrinks as weight mass
concentrates.

% ———————————————————————
% 6. Future-horizon discrimination
% ———————————————————————
\subsection*{Future-horizon discrimination}

A central question for a trajectory model is not merely what a patient’s
next visit will contain, but how far into the future a diagnosis can be
anticipated. We measure this with a \emph{future-horizon} analysis: at
each scored \texttt{VISIT\_START} position, we record the horizon from
that visit boundary to the \emph{nearest future occurrence} of the target
code anywhere in the remainder of that patient’s trajectory. Scored
positions are then assigned to one of six lead-time bands:
\textit{Same day}, \textit{\textless1mo}, \textit{1–6mo},
\textit{6–12mo}, \textit{1–2y}, and \textit{\textgreater2y}
(Table~\ref{tab:horizon-bins}). For each ICD-10 code and each horizon
band, we compute a first-occurrence AUC comparing cases whose next
occurrence of the target code falls in that band against matched controls
who never receive the target code. This analysis measures how rapidly
diagnostic discrimination decays as the forecasting window lengthens.

Each per-code, per-horizon AUC reuses the case/control estimator and the
sex $\times$ age matching cascade described above. Controls are drawn from
patients who do not carry the target code anywhere in their full
trajectory, and are reweighted to match the sex and age distribution of
cases within each code–horizon stratum. We require at least 100 total
future cases per code, at least five cases in a code–horizon cell, at
least 20 eligible controls, and a minimum matching coverage of 0.5. The unit of analysis is the scored position, not the patient: for each code, every eligible pre-diagnosis \texttt{VISIT\_START} position
contributes a separate case to the horizon band matching its own distance to that code's nearest future occurrence, so a single patient 
may contribute multiple case observations to the same code — even within a single horizon band, if several of their visits precede the 
same eventual diagnosis at similar lags — and the patients and codes contributing to the estimate may vary across lead times.

To test whether the temporal decline is driven by changing cohort
composition rather than loss of predictive signal, we also run a
fixed-cohort sensitivity analysis. For each code, we restrict to patients
who are evaluable as first-occurrence cases in every active horizon band
for that code, and then compute the same matched per-code AUC separately
within each band. Thus, within a code, the same case set is followed
across lead times. This fixed-cohort analysis is more stringent and
retains fewer codes and observations, but it makes the horizon trend less
sensitive to differences in which patients contribute to each bin.

For both the unrestricted and fixed-cohort analyses, the figure reports
the distribution of per-code AUCs across conditions within each horizon
band. Supplementary Table~\ref{tab:future-horizon} reports the median
per-code AUC and a 95\% bootstrap confidence interval for that median,
computed with 2{,}000 condition-level resamples.

\begin{table}[h]
\centering
\small
\caption{Future-horizon bins. The horizon is the time from a scored
\texttt{VISIT\_START} position to the nearest future first occurrence of
the target code. Bins are defined by lead time to first occurrence.}
\label{tab:horizon-bins}

\begin{tabular}{ll}
\toprule
Bin & Lead time to first occurrence \\
\midrule
Same day & 0 days \\
$<$1mo    & $>0$ to $<30$ days \\
1--6mo    & 30 to $<180$ days \\
6--12mo   & 180 to $<365$ days \\
1--2y     & 365 to $<730$ days \\
$>$2y     & $\geq 730$ days \\
\bottomrule
\end{tabular}

\end{table}

% ---------------------------------------------------------------------
% 7. Stratified AUC: sex, age, and prevalence                   [POLISH]
% ---------------------------------------------------------------------
\subsection*{Stratified AUC: sex, age, and prevalence}

% =====================================================================
% KEY FACTS (origin/main, 2026-06-02)
%   Source: teddy/eval/_shared/auc_stratified.py (by_sex / by_age),
%           scripts/auc_rare_vs_common.py (prevalence split).
%   - by-sex: per-code AUC within female / male strata; unknown dropped.
%     Writes per_code_auc/by_sex/per_code_auc_by_sex.csv.
%   - by-age: per-code AUC within the 5-band AAP scheme
%     (0-1y,1-3y,3-6y,6-12y,12-18y+); out-of-band positions dropped.
%     Writes per_code_auc/by_age/per_code_auc_by_age.csv.
%   - rare-vs-common: single participant-prevalence threshold (default
%     5/10,000 = 0.0005). code is "rare" if prevalence < threshold,
%     else "common". Same Mann-Whitney estimator within each stratum.
%   - All three reuse the section-5 estimator (one-rep-per-patient
%     control sampling), so they are directly comparable to the pooled
%     per-code AUC column.
% =====================================================================

Beyond the pooled per-code AUC, we report three stratified views
that surface where discrimination varies across the cohort. Each view
applies the same Mann--Whitney estimator and one-position-per-patient
control sampling as in the Per-code AUC section, computed independently within each stratum, so
the stratified AUCs are directly comparable to the pooled column. First,
we stratify by sex, computing per-code AUC separately within the
female and male strata (positions of unknown sex are excluded). Second,
we stratify by age using the 5-band AAP scheme (0--1y, 1--3y,
3--6y, 6--12y, 12--18y+), with positions outside all bands excluded;
this reveals how a code's predictability shifts across the pediatric age
range. Third, we stratify by prevalence, splitting the
vocabulary into \emph{rare} and \emph{common} codes at a single
participant-prevalence threshold (a code is rare if fewer than five in
ten thousand patients ever receive it) and comparing the AUC
distributions of the two strata, which quantifies how much of the
model's discrimination is carried by common versus long-tail codes.

% ---------------------------------------------------------------------
% 8. Temporal calibration of time-to-next-visit                  [POLISH]
% ---------------------------------------------------------------------
\subsection*{Temporal calibration of time-to-next-visit}

% =====================================================================
% Backs Results section 4 (Fig. 5). Numbers from results.tex / fig5:
%   n = 100,000 VISIT_START positions (held-out test split).
%   median observed inter-visit interval 21 d vs predicted 30 d.
%   log-scale correlation approx 0.40.
%   RMST (restricted mean survival time) mean error 2.7 d.
%   S(t) = P(no visit yet): Teddy exp(-lambda t) vs Kaplan-Meier.
% Time is supervised only at visit boundaries (time_loss_on_visit_only,
% section 3), so calibration is read where supervision applies.
% =====================================================================

Because \textsc{TEDDY} supervises elapsed time only at visit boundaries and \texttt{NO\_EVENT} checkpoints (\S3), we assess the calibration
of its predicted time-to-next-visit at exactly those positions. We draw a sample of $200{,}000$ scored positions from the held-out test
split whose target is \texttt{VISIT\_START} (an observed visit) or \texttt{NO\_EVENT} (a right-censored inter-visit checkpoint), and at
each, convert the time head's predicted rate $\lambda_t$ into a survival function $S(t)=\exp(-\lambda_t t)$, the model's probability that
no further visit has occurred by elapsed time $t$. We compare the population-averaged model survival curve against the empirical
Kaplan--Meier estimate over the observed inter-visit intervals, treating \texttt{VISIT\_START} as an event and \texttt{NO\_EVENT} as
censoring. To summarise agreement on the restricted-mean-survival-time (RMST) scale, we group the sampled positions into ten deciles of
predicted RMST (area under the model's survival curve out to a 365-day horizon), and within each decile compare the mean predicted RMST
against the observed RMST (the area under the decile's KM curve). The reported mean absolute RMST error is the sample-size-weighted
average of this predicted-vs-observed gap across the ten deciles, not an average over individual positions. Because the head
emits a single exponential rate per position, this evaluation probes
whether that rate is context-dependent and well-scaled rather than
whether the inter-visit distribution is itself exponential; any
divergence of the fitted exponential from the Kaplan--Meier tail
therefore reflects the deliberate simplicity of the per-position model
rather than a miscalibration of its central tendency.

\subsection*{Baselines}

We compare \textsc{TEDDY} against two baselines, each calibrated to a
different counterfactual.

% \textbf{ICD-prior logistic regression.} For each ICD code in the
% vocabulary we fit a binary logistic regression on three feature blocks:
% a one-hot vector of ICD codes occurring in the patient's past-window
% (a fixed look-back of recent days), a three-level sex one-hot, and a
% six-level age-band one-hot. Per-code AUC is computed on the same test
% split and reported at the same future-horizon bins as \textsc{TEDDY}, via the
% baseline's matched future-horizon mode (which consumes \textsc{TEDDY}'s matched
% control cache so the case/control construction is identical). This
% baseline establishes
% a demographic and recent-history floor: the gap between \textsc{TEDDY}'s AUC
% and the ICD-prior LR AUC is, by construction, what the transformer
% learns \emph{beyond} sex, age, and past-window co-occurrence.

\textbf{Alternative sequence architectures.} To isolate the effect of
the transformer architecture from the effect of training a generative
model on EHR sequences, we train four alternative architectures --- a
two-layer LSTM, a two-layer vanilla RNN, a two-layer CNN
(kernel size 3), and a two-layer DenseNet --- on the same canonical
dataset, with matching embedding dimension ($n_{\mathrm{embd}}=128$),
context window (2{,}048 tokens), optimiser, learning-rate schedule, and
mixed-precision settings. These baselines are size-matched to
\textsc{TEDDY} in width and training recipe and are no larger in
parameter count, so the only model in our comparison that exceeds
\textsc{TEDDY}'s scale is the general-purpose Gemma-3-4b introduced below.
All four are evaluated through the same per-code AUC pipeline as
\textsc{TEDDY}.
% VERIFY (§10 baselines): the local lstm_baseline.yaml shows n_embd=120 on
% the OLD non-bracketed data. Confirm (a) the baseline width matches
% TEDDY's headline n_embd=128 (else soften "matching embedding dimension"/
% "size-matched"), and (b) all four baselines were retrained on the
% visit-bracketed 1,849-vocab dataset, so the per-code AUC comparison is
% on the same data as the headline model.

\textbf{General-purpose LLM head-to-head.} As an external comparator
we benchmark \texttt{google/gemma-3-4b-it}, a 4-billion-parameter
instruction-tuned LLM accessed via Amazon Bedrock, on a binary
next-ICD-code task framed identically for both models so the results
are directly comparable. For a chosen target code (J45 / asthma is the
default headline; the cohort builder accepts any 3-character ICD-10
code), we draw the case/control cohort from the same test split as
\textsc{TEDDY}'s evaluation. Cases are patients with at least one recorded
occurrence of the target code; the prediction cutoff is set at the
first occurrence, with at least five real ICD events required in the
preceding history. Controls are patients with no occurrences of the
target and at least one ICD event after a sampled cutoff; the cutoff
position is drawn from the cases' empirical history-length CDF
(seed 42) so length cannot trivially separate the two groups. We
optionally enforce strict length matching (Kolmogorov-Smirnov $\approx 0$)
as a sensitivity check.

Each patient's pre-cutoff history is rendered as a single text prompt
under one of four variants that vary along two orthogonal axes ---
representation (raw ICD codes vs.\ ICD-10 subblock terms) and ordering
(top-frequency unordered vs.\ chronological with age annotations) ---
giving \texttt{codes\_only}, \texttt{code\_at\_age},
\texttt{terms\_only}, and \texttt{terms\_at\_age}. The model is asked to
return a single floating-point risk score, which is parsed by regex
from the response. We use \texttt{temperature}~$=0$, max-new-tokens 64,
4-way thread parallelism, and exponential-backoff retries on transient
Bedrock errors (up to 5 attempts).

To put \textsc{TEDDY} on the same scoring axis we exploit the
competing-exponentials structure of its head: under that
parametrisation, $\mathrm{softmax}(\mathbf{z}_t)_c$ at the last valid
position is exactly the probability that the next event is code $c$.
We slice each cohort patient's trajectory to the same cutoff, run a
single forward pass through the canonical \textsc{TEDDY} checkpoint, and read
out $P(\text{next}=\text{target})$ directly from the output softmax.
Both models are scored on AUROC and average precision over the same
cohort, so any performance difference between Gemma and \textsc{TEDDY} can be
attributed to the modelling and not to differences in cohort
construction or scoring axis.

\subsection*{Reproducibility and artifacts}

The patient-level electronic health record data underlying this study
contain protected health information and \emph{cannot} be shared: neither
the raw records nor the de-identified tokenized corpus, trained model
checkpoints, or per-position evaluation caches are publicly released, as
each is derived from protected patient data and is subject to the
institutional review board approval and data-use agreements governing the
TCH EHR. Access to derived
artifacts may be considered on reasonable request to the corresponding
authors, subject to institutional approval and an appropriate data-use
agreement. The implementation, configuration files, and analysis scripts
that produce every result in this paper are open source at
\url{https://github.com/LiuzLab/teddy}, so the full pipeline can be
reproduced against an equivalently formatted cohort.

\section*{Ethics Approval}

This study was conducted in accordance with the Declaration of Helsinki
and approved by the Baylor College of Medicine Institutional Review Board
(protocol code H-52222; date of approval 2 October 2025). The requirement
for informed consent and HIPAA authorization was waived because the study
involved retrospective analysis of existing clinical data, posed no greater
than minimal risk to participants, and obtaining individual consent was
impracticable.

\bibliographystyle{unsrt}  

\bibliography{references}

% ---------------------------------------------------------------------
% Placeholder labels for supplementary items not yet created
% ---------------------------------------------------------------------
\clearpage
\appendix
\setcounter{figure}{0}
\renewcommand{\thefigure}{S\arabic{figure}}
\setcounter{table}{0}
\renewcommand{\thetable}{S\arabic{table}}

\section*{Supplementary Figures and Tables}

% Rare-code AUC confidence intervals (supports Results section 1)
\begin{figure}[h]
\centering
\includegraphics[width=0.8\linewidth]{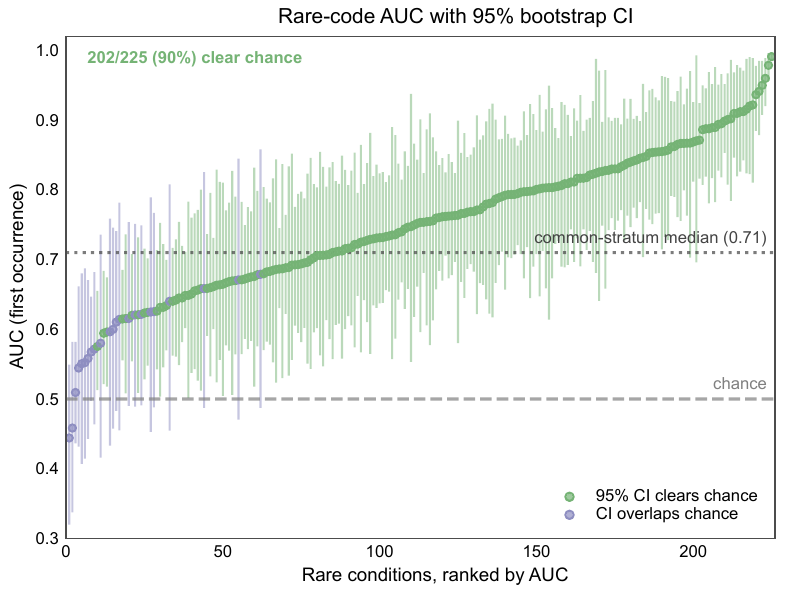}
\caption{\textbf{Rare-condition discrimination is signal, not
small-sample scatter.} Each of the 225 rarest conditions (participant
prevalence $<5\times10^{-4}$) is shown with its first-occurrence AUC and
a 95\% confidence interval from a patient-level bootstrap
(2{,}000 resamples), using the same sex- and age-matched estimator as
the main text. Conditions are ranked by AUC along the horizontal axis so
every interval is legible; points are coloured by whether the CI clears
chance. The dashed line marks chance (AUC${=}0.5$) and the dotted line
the common-stratum median (0.71). 202 of the 225 rare conditions (90\%)
have a CI lying entirely above chance, confirming that the wider AUC
scatter at low prevalence is an expected small-sample effect rather than
absence of signal.}
\label{fig:rare-ci}
\end{figure}
\addcontentsline{toc}{section}{Supplementary Figures and Tables}

\begin{figure}[h]
\centering
\includegraphics[width=\linewidth]
{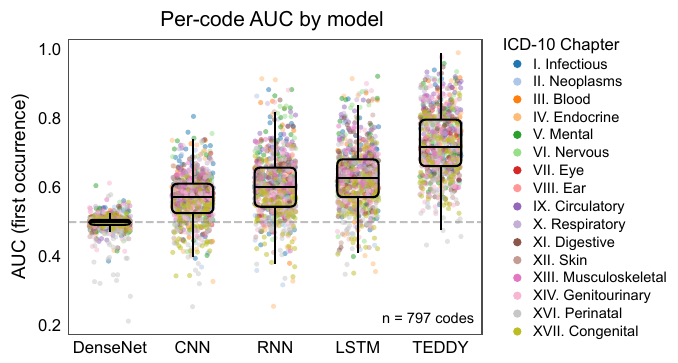}
\caption{\textbf{\textsc{TEDDY} outperforms alternative architectures
across 797 incident-diagnosis tasks.} Per-code first-occurrence AUC
distributions are shown for DenseNet, convolutional neural network (CNN),
recurrent neural network (RNN), long short-term memory (LSTM), and
\textsc{TEDDY} models evaluated on the held-out test set. Each point
represents one ICD-10 diagnosis code and is colored by ICD-10 chapter;
boxplots summarize the distribution across the 797 eligible conditions
from the 16 disease-bearing chapters analyzed in the main text. All
models are evaluated using the same visit-boundary protocol, in which
diagnosis risk is scored before any codes from the target visit are
visible. AUC is calculated for first recorded occurrences using
sex- and age-matched controls, as described in Methods. The dashed
horizontal line marks chance discrimination (AUC${=}0.5$).}
\label{fig:benchmark_all_codes}
\end{figure}

\begin{table}[h]
\centering
\caption{\textbf{Paired comparison of per-code first-occurrence AUC
between \textsc{TEDDY} and same-data baseline architectures.}
All architectures were evaluated on the same 797 ICD-10 codes with
defined first-occurrence AUCs, so each comparison is paired by diagnosis.
$\Delta$AUC is defined as
AUC$_{\textsc{TEDDY}}-$AUC$_{\text{comparator}}$; positive values favor
\textsc{TEDDY}. $P$ values are from two-sided paired Wilcoxon signed-rank
tests and are Holm-adjusted across the four comparisons.}
\begin{tabular}{lcccc}
\toprule
Comparator & Median AUC & Median $\Delta$AUC &
Codes favoring \textsc{TEDDY} & Adjusted $P$ \\
\midrule
DenseNet & 0.500 & 0.224 & 792/797 (99\%) & $3.9\times10^{-131}$ \\
CNN      & 0.572 & 0.158 & 784/797 (98\%) & $5.0\times10^{-130}$ \\
RNN      & 0.601 & 0.117 & 775/797 (97\%) & $1.7\times10^{-129}$ \\
LSTM     & 0.627 & 0.095 & 769/797 (96\%) & $5.0\times10^{-126}$ \\
\bottomrule
\end{tabular}
\label{tab:paired_model_benchmark}
\end{table}

\begin{figure}[h]
\centering
\includegraphics[width=\linewidth]{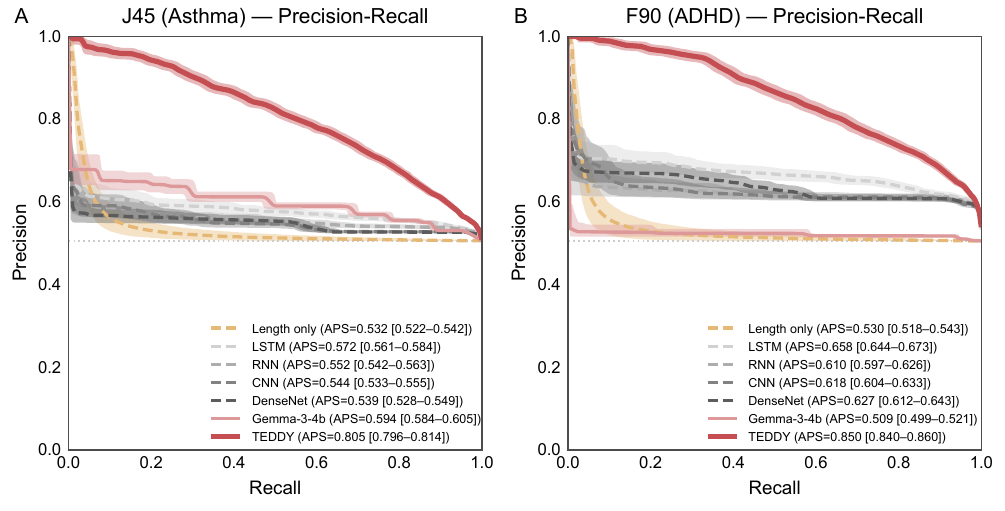}
\caption{\textbf{\textsc{TEDDY} achieves the highest average precision
for incident asthma and ADHD prediction.} Precision–recall curves are
shown for the same head-to-head cohorts used in Fig.~\ref{fig:benchmark}:
\textbf{(A)} asthma (J45) and \textbf{(B)} attention-deficit/hyperactivity
disorder (ADHD; F90). Cases are children scored immediately before the
visit containing their first recorded target diagnosis, and controls are
children who never receive that diagnosis, with control cutoffs sampled
to match the case history-length distribution. Average precision
summarizes how well each model enriches children who later develop the
target condition. \textsc{TEDDY} achieves the highest average precision
for both targets: 0.805 (95% CI 0.796–0.814) for asthma and 0.850
(0.840–0.860) for ADHD. For asthma, the strongest non-\textsc{TEDDY}
comparator is Gemma-3-4b at 0.594 (0.584–0.605), and the strongest
same-data sequence baseline is the LSTM at 0.572 (0.561–0.584). For
ADHD, the strongest non-\textsc{TEDDY} comparator is the LSTM at 0.658
(0.644–0.673). Shaded bands and bracketed legend values denote 95%
patient-level bootstrap confidence intervals from 2{,}000 resamples.}
\label{fig:benchmark_pr}
\end{figure}

% % Placeholder: dyad OOV table
% \begin{table}[h]
%   \centering
%   \begin{tabular}{lll}
%     \toprule
%     Code & Source & Occurrences \\
%     \midrule
%     O11 & Maternal & 4{,}639 \\
%     O12 & Maternal & 2{,}575 \\
%     O22 & Maternal & 2{,}523 \\
%     \multicolumn{3}{l}{\textit{(full list in supplementary)}} \\
%     \bottomrule
%   \end{tabular}
%   \caption{Out-of-vocabulary codes in the dyad fine-tune cohort.}
%   \label{tab:dyad-oov}
% \end{table}

\begin{table*}[h]
\centering
\caption{Future-horizon first-occurrence AUC by lead time to first occurrence. \textbf{A. No cohort filtering}: every first-occurrence case per horizon band, all eligible ICD-10 codes. \textbf{B. Fixed cohort}: for each code, patients evaluable as a first-occurrence case in every active band, so each band scores the same patient set per code. Per-code AUC is weighted Mann--Whitney with sex- and age-matched controls; the median and a 95\% bootstrap CI (2{,}000 condition-level resamples) summarise the per-code AUC distribution within each band. $n_{\text{patients}}$ is distinct patients contributing (union over codes); $n_{\text{obs}}$ is per-code first-occurrence case observations at the scored-position level; a patient may contribute more than one 
observation per code (one per eligible pre-diagnosis visit), including more than one within the same horizon band.}
\label{tab:future-horizon}
\begin{tabular}{l rr r r r rr r r r}
\toprule
 & \multicolumn{5}{c}{\textbf{A. No cohort filtering}} & \multicolumn{5}{c}{\textbf{B. Fixed cohort}} \\
\cmidrule(lr){2-6}\cmidrule(lr){7-11}
Horizon & Median & 95\% CI & $n_{\text{codes}}$ & $n_{\text{patients}}$ & $n_{\text{obs}}$ & Median & 95\% CI & $n_{\text{codes}}$ & $n_{\text{patients}}$ & $n_{\text{obs}}$ \\
\midrule
Same day & 0.718 & [0.711, 0.730] & 797 & 74,452 & 556,972 & 0.869 & [0.856, 0.880] & 578 & 14,515 & 54,254 \\
$<$1\,mo & 0.761 & [0.750, 0.771] & 675 & 41,956 & 205,217 & 0.832 & [0.821, 0.843] & 475 & 13,948 & 52,787 \\
1--6\,mo & 0.704 & [0.696, 0.714] & 675 & 48,597 & 336,607 & 0.794 & [0.781, 0.806] & 461 & 13,757 & 52,409 \\
6--12\,mo & 0.675 & [0.667, 0.684] & 623 & 43,426 & 287,635 & 0.760 & [0.750, 0.771] & 436 & 13,372 & 51,892 \\
1--2\,y & 0.636 & [0.624, 0.646] & 618 & 42,395 & 270,013 & 0.730 & [0.720, 0.741] & 438 & 13,336 & 51,915 \\
$>$2\,y & 0.597 & [0.589, 0.607] & 608 & 39,875 & 223,632 & 0.644 & [0.635, 0.652] & 437 & 13,324 & 51,903 \\
\bottomrule
\end{tabular}
\end{table*}

\end{document}